\documentclass[letterpaper, 10 pt, conference]{ieeeconf}  

\IEEEoverridecommandlockouts                              

\usepackage{graphicx}
\usepackage{caption}
\usepackage{amsmath}
\usepackage{amssymb}
\usepackage{hyperref}
\usepackage{amsmath}
\usepackage[linesnumbered,ruled]{algorithm2e}
\usepackage{tabularx}
\usepackage{cite}

\title{\LARGE \bf
NanoMap:  Fast, Uncertainty-Aware Proximity Queries with \\ Lazy Search over Local 3D Data
}

\author{Peter R. Florence$^{1}$, John Carter$^{1}$, Jake Ware$^{1}$,  Russ Tedrake$^{1}$
\thanks{$^{1}$CSAIL, Massachusetts Institute of Technology, Cambridge, MA, USA
        {\tt\small \{peteflo,jcarter,jakeware,russt\}@csail.mit.edu}}%
}

\begin{document}

\maketitle
\thispagestyle{empty}
\pagestyle{empty}

\begin{abstract}
We would like robots to be able to safely navigate at high speed, efficiently use local 3D information, and robustly plan motions that consider pose uncertainty of measurements in a local map structure. This is hard to do with previously existing mapping approaches, like occupancy grids, that are focused on incrementally fusing 3D data into a common world frame.  In particular, both their fragile sensitivity to state estimation errors and computational  cost can be limiting.  We develop an alternative framework, NanoMap, which alleviates the need for global map fusion and enables a motion planner to efficiently query pose-uncertainty-aware local 3D geometric information.  The key idea of NanoMap is to store a history of noisy relative pose transforms and search over a corresponding set of depth sensor measurements for the minimum-uncertainty view of a queried point in space.  This approach affords a variety of capabilities not offered by traditional mapping techniques: (a) the pose uncertainty associated with 3D data can be incorporated in motion planning, (b) poses can be updated (i.e., from loop closures) with minimal computational effort, and (c) 3D data can be fused lazily for the purpose of planning.  We provide an open-source implementation of NanoMap, and analyze its capabilities and computational efficiency in simulation experiments.  Finally, we demonstrate in hardware its effectiveness for fast 3D obstacle avoidance onboard a quadrotor flying up to 10 m/s.
\end{abstract}

\section{INTRODUCTION}

Robust, fast motion near obstacles is an open problem that is central in robotics, with applications spanning across manipulation, autonomous cars, and UAV navigation in unknown environments.  Although many approaches exist for planning obstacle-free motions, mapping errors due to significant state estimation uncertainty can degrade their performance \cite{liu2016high, florence16}.  Accordingly, a notable trend in the state of the art has been to develop memoryless approaches to obstacle avoidance that use only the current depth sensor measurement \cite{matthies2014stereo, liu2016high, daftry2016robust, florence16, lopez17}.  These approaches are less prone to state estimation errors, but fail to capture all available information.

\begin{figure}[htbp]
\hspace*{-0.2cm}                                                           
   \includegraphics[scale=0.55]{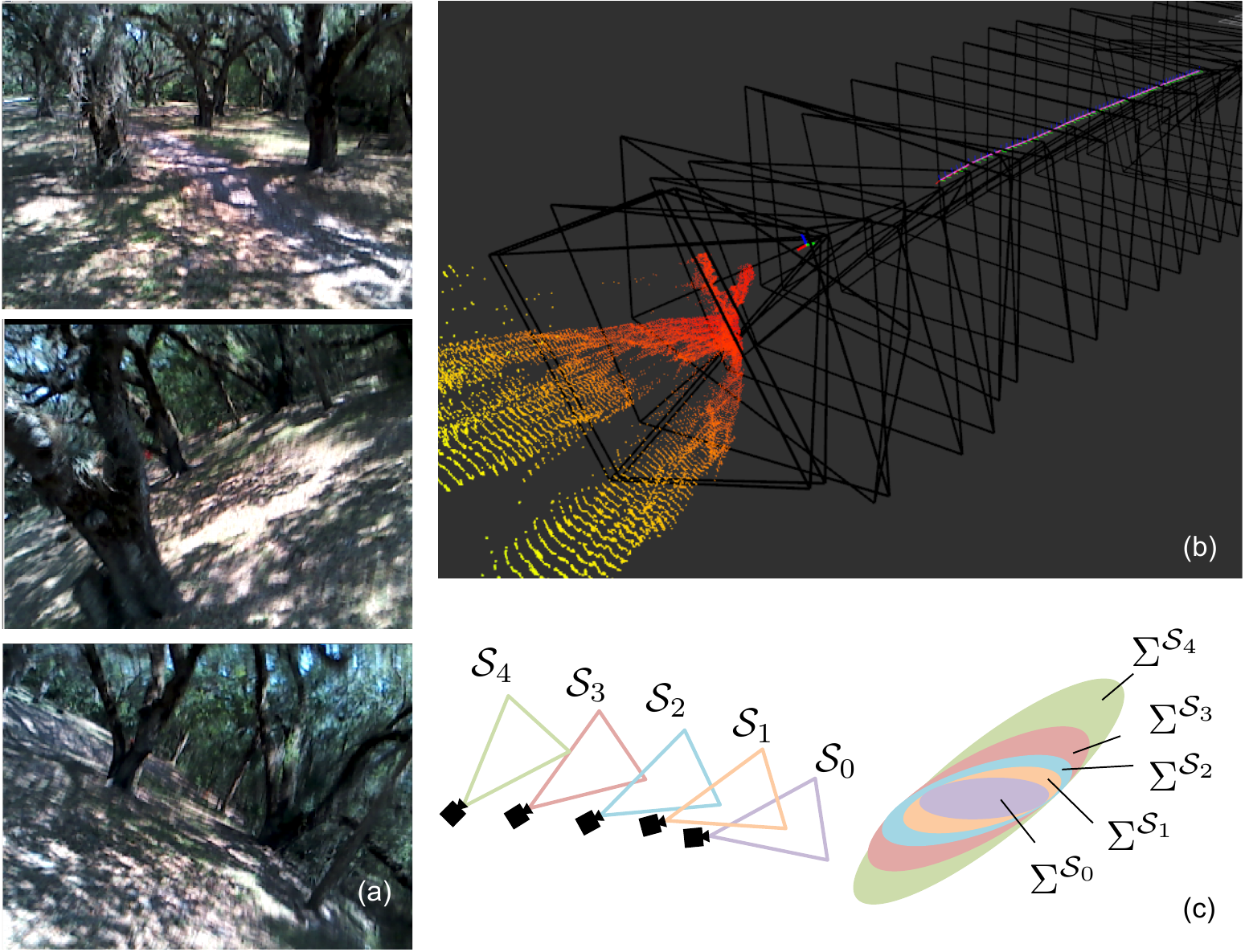}
  \caption{(a) Onboard images from a quadrotor using NanoMap and flying at 10 $m/s$ in a forest.  (b) Visualization of vehicle's depth camera frustums over time, and current point cloud observing a tree.  (c) Depiction of frame-specific uncertainty $\Sigma^{\mathcal{S}_i}$ for each depth sensor measurement frame $\mathcal{S}_i$.}
  \label{intro_figure}
\end{figure}

Towards this goal, a primary motivation of this work was to be able to use pose uncertainty to reason about a local history of depth information.  NanoMap is an algorithm and data structure that enables uncertainty-aware proximity queries for planning.  While traditional mapping approaches rely on fusing a history of depth information into a discretized world frame, we propose an alternative: perform no discretization, and no fusing.  Instead, the process for querying local 3D data is a \textit{search over views}.  When a query point (i.e. a sample along a motion plan) is provided, the history of depth information is searched for the most-recent and therefore minimum-uncertainty relative to current body frame view of that query point.

In practice, this approach offers a variety of unique capabilities not present in traditional fusion-based mapping algorithms.  For one, the pose uncertainty associated with depth sensor measurements can be incorporated into planning, by treating each pose with frame-specific uncertainty relative to the current body frame (Figure 1, c).  Second, since fusion between measurements is not performed, it is trivial to incorporate updated information about previous poses.  Third, the build time of the data structure is low, which leads to an improvement in computational efficiency for small amounts of motion planning queries ($< 10,000$).

This paper presents the design of NanoMap and our experiments in quantifying the benefits of its novel properties.  
We believe this work strongly demonstrates that more deeply integrating motion planning and perception can improve a system's robustness and computational efficiency.  To briefly clarify our scope of work: (a) we focus on a method of incorporating pose uncertainty, but modeling the noise of the depth sensor itself is outside of scope, (b) NanoMap requires nonzero volume depth sensors, i.e. depth cameras or 3D lidars, but not 2D or 1D sensors, (c) adding more sensors to increase the FOV is a hardware route to alleviate the problem but does not address occlusions, and (d) we are concerned with local obstacle avoidance, rather than global planning, and so short histories of information are sufficient.  

The contributions of this work are as follows:
\begin{itemize}
\item A novel use of frame-specific uncertainty for planning with depth sensors
\item An approach to searching a history of depth frustums to enable motion plans to satisfy field of view constraints
\item An efficient use of independently spatially partitioned depth measurements for motion planning queries
\item Simulation experiments demonstrating the magnitudes of state estimation uncertainty at which frame-specific uncertainty becomes significant (approximately $1\%$ drift, or 1 $m$ pose corrections)
\item Hardware validation demonstrating this approach onboard a quadrotor, including flight at up to $8-10$ $m/s$ in unknown warehouse and forest environments
\end{itemize}

\section{RELATED WORK}


A few related works share some features of using pose estimation uncertainty in planning, but do not address planning around obstacles in unknown environments.  Previous works have used directly the uncertainty of a pose graph framework for planning but have a critical limitation that they only plan over graphs of pre-known poses \cite{valencia2013planning, teniente2011dense}.  Other work seeks to develop generalized belief space that includes distributions over worlds, but there are no obstacles in these worlds, only landmarks for navigation \cite{indelman2016towards}.  Another related work includes a sampling of depth perception estimates (a  discrete probability distribution), but inserts them into a map structure using maximum-likelihood poses \cite{dey2016vision}.

Rather than deal with the belief space of previous poses, the predominant approach for incorporating memory has been to ignore pose uncertainty, and use a maximum-likelihood mapping approach \cite{burri2015real,oley16}. Mapping-based approaches benefit from extensive decades of research into the robot mapping problem.  While many SLAM approaches may internally have rich representations of uncertainty from the fusion of a variety of noisy depth sensor, RGB, and other sensor data, when it comes to using maps for planning, the maximum likelihood estimate map $\mathcal{\hat{{M}}}_{MLE}$ is traditionally used.  There are a variety of different ways to formulate a map -- the most common versions are occupancy grids, which are used ubiquitously \cite{Hornung:2013:OEP:2458738.2458796}.  Occupancy grids can probabilistically incorporate depth sensor measurements (multiple measurements can be required for a cell to be occupied), but this doesn't address pose estimation uncertainty.  Other forms include polar maps, and for some dense SLAM techniques, surfel maps are used.  
Some probabilistic collision detection methods can also handle non-spherical robots \cite{pan2017probabilistic} and dynamic obstacles \cite{park2016fast}, whereas we have only used NanoMap here with a method \cite{florence16} that assumes spherical robot and static environment.   

A different and popular approach to the obstacle avoidance problem under significant state estimation uncertainty is to essentially cut pose estimation out of the equation, which can be done via a method that uses no memory of depth sensor measurements.  In addition to planning-based approaches that exhibit this property \cite{otte2009path, florence16, daftry2016robust, matthies2014stereo, liu2016high, lopez17}, any obstacle avoidance approaches that are considered reactive approaches may inherently have this property as well.  Reactive approaches, including optic flow methods \cite{beyeler2009vision}, 
reactive imitation-learning \cite{ross2013learning}, and non-planning-based geometric approaches \cite{oleynikova2015reactive} have demonstrated considerable success at obstacle avoidance for UAVs.   The limitations, however, of memoryless obstacle avoidance have been well noted \cite{otte2009path, ross2013learning}, including that the restricted memoryless free space provides less space for dynamic manuevers.  Other related approaches have limited map-building to very short time horizons \cite{barry2017}, or have used map structures that exponentially decay old depth sensor measurements \cite{dey2016vision}.

\section{MOTIVATION}

This work seeks a method to reason about local 3D obstacles in the presence of significant state estimation uncertainty.  Our approach is guided by our experience with high-speed UAVs, the use of depth sensors for obstacle avoidance, and the planning challenges introduced by imperfect state estimation \cite{florence16}.

\begin{figure}
  \includegraphics[keepaspectratio=true,scale=0.59]{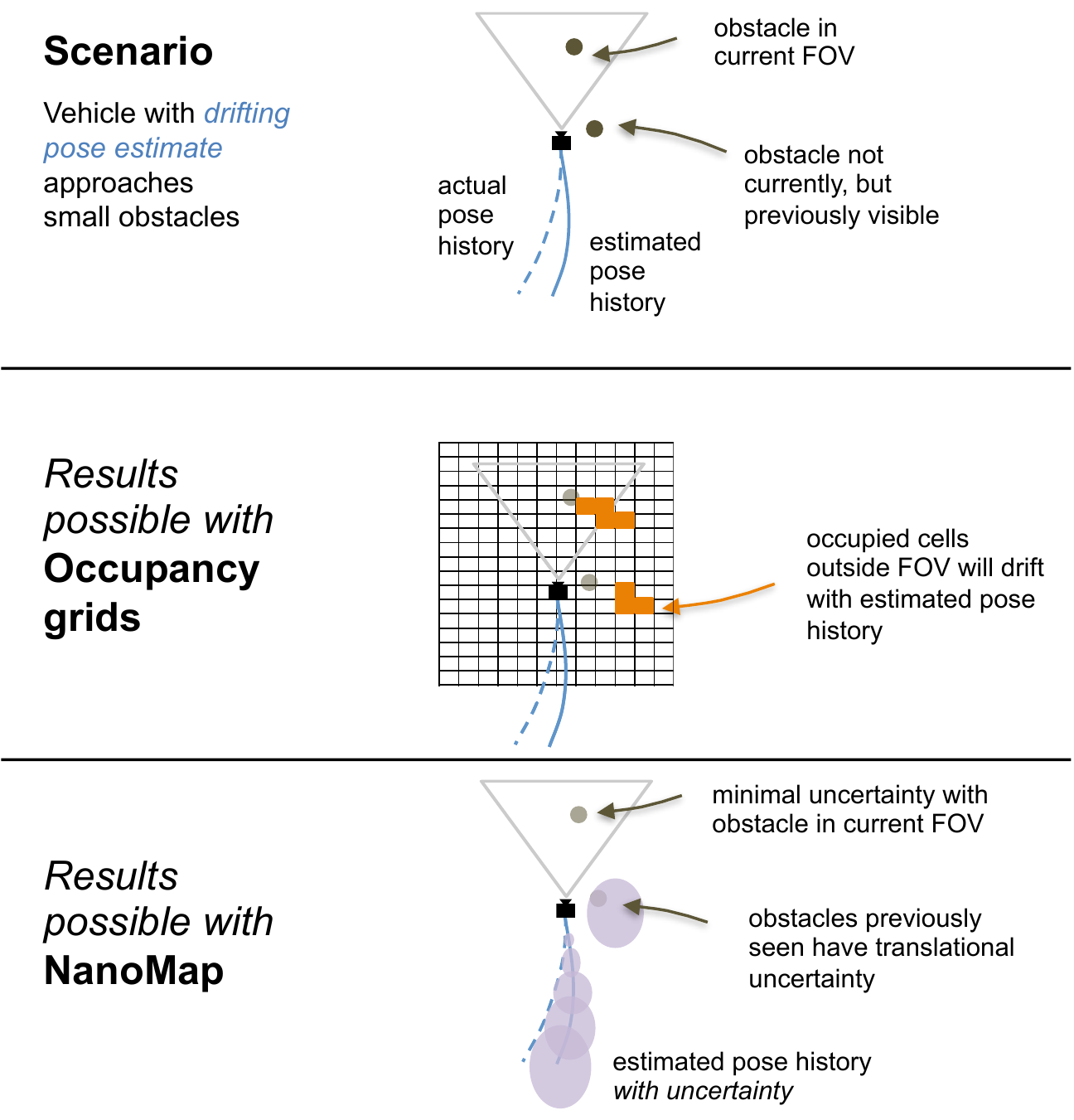}
  \captionof{figure}{Comparison of possible results using an occupancy grid (middle), vs. NanoMap (bottom), in an example scenario navigating amongst obstacles while experiencing pose drift.}\label{occ_grid_vs_nanomap}
\end{figure}

One key observation is that in practice, depth sensor data (Figure 1, b) is often clean enough that fusing many recent observations is not required in order to plan obstacle-free motions.  Rather than averaging many measurements to create intricate 3D reconstructions, mapping for obstacle avoidance only needs to robustly determine collision-free space. Furthermore, the current or very recent depth measurements frequently contain a view of planned directions of motion (Figure 1, b).  In the case that the planned trajectory does not fall within the current field of view, it is still possible to perform robust trajectory planning by using the history of depth measurements. 

Additionally, as shown in Figure \ref{occ_grid_vs_nanomap}, the incorporation of pose uncertainty (the acknowledgement that the robot does not perfectly know its previous positions relative to its current body frame) is a fundamentally different model of uncertainty than, for example, what is modeled in an occupancy grid.  Although the Bayesian update in occupancy grids may well model 0-mean Gaussian noise of both poses and depth sensing, it does not handle the case of pose drift.

\section{FORMULATION}

\begin{figure}
  \includegraphics[keepaspectratio=true,scale=0.4]{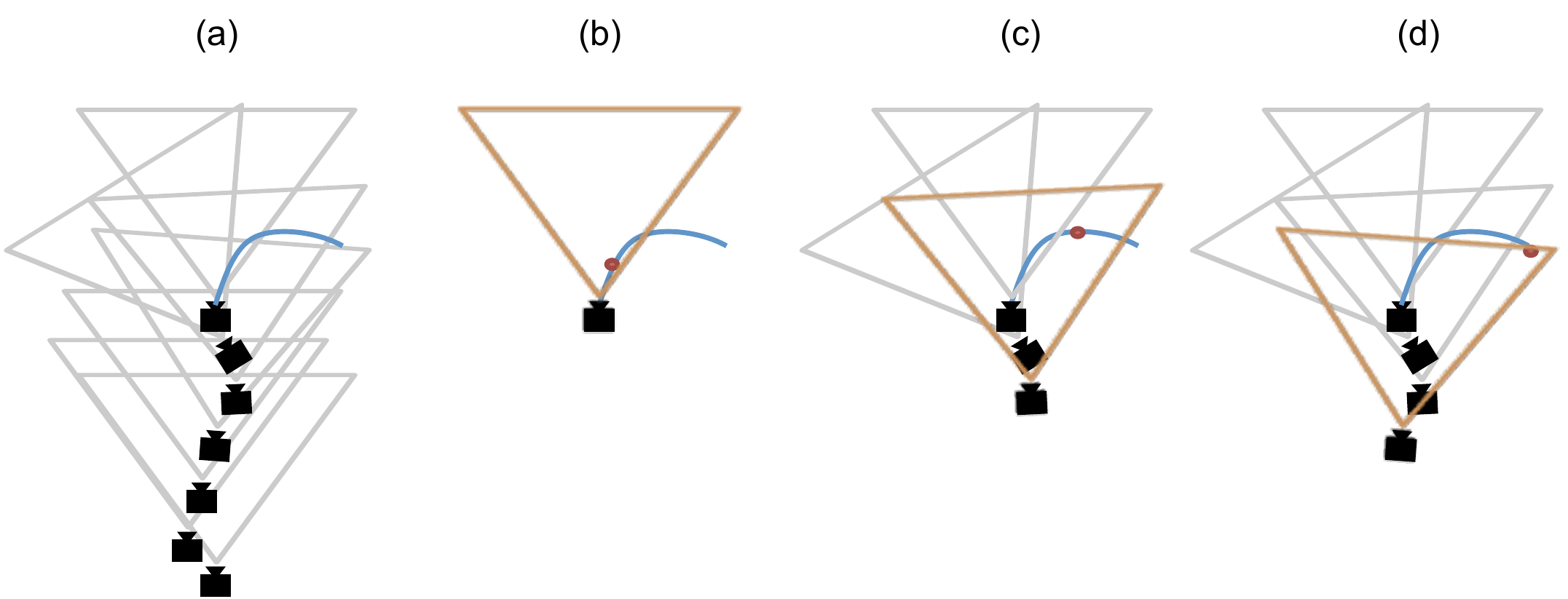}
  \captionof{figure}{Depiction of how NanoMap queries can be used to evaluate motion plans, (a, blue), given a series of depth sensor measurements over time (a, camera frustums). For each sample point (b, c, d; red) the history of measurements is searched until a view is found (orange) that contains the sample point.  Note that sample points are actually Gaussian sample points (See Figure \ref{whichfov}).}\label{reverse_search}
\end{figure}

NanoMap is a framework composed of both a local 3D data structure and an algorithm for searching that data structure.  Briefly, the algorithm works by reverse searching over time through sensor measurement views until finding a satisfactory view of a subset of space (Figure \ref{reverse_search}), and then returning the $k$-nearest-neighbors from that view's sensor measurement.  Important components of the framework include: the determination of in-frame views (the IsInFOV() function), the propagation of uncertainty, and efficient data structure design for handling asynchronous data inputs of point clouds, poses, and pose updates. We first describe the query algorithm, which gives insight into efficient data structure design.  We then discuss details of handling asynchronous data.

\subsection{Querying Algorithm}

The query algorithm (Algorithm \ref{nmknn}) iteratively transforms an uncertain query point into the coordinate frames of previous sensor measurements until it finds a view which contains the query.  An uncertain query point is a sampled point along a stochastic motion plan, and is provided in body frame, $\mathbf{x}^{\mathcal{B}}_{query} = \mathcal{N}(\mathbf{\mu}^{\mathcal{B}}, \Sigma ^{\mathcal{B}} ) \in \mathbb{R}^3$. The query point in the original body frame and each of the relative transforms are each modeled with Gaussian translational uncertainty.  In each frame associated with a given sensor measurement $\mathcal{S}_{i}$, the query point $\mathbf{x}^{\mathcal{S}_{i}}_{query} = \mathcal{N}(\mathbf{\mu}^{\mathcal{S}_{i}}, \Sigma ^{\mathcal{S}_{i}} ) \in \mathbb{R}^3$ has uncertainty specific to that frame.  As noted in Algorithm \ref{nmknn}, NanoMap is unconventional in that it also returns the uncertain query point itself transformed into a different frame. While NanoMap has been implemented to only address query points in $ \mathbb{R}^{3} $, downstream the query return points may be inflated for spherical approximations of collision geometry.

\begin{algorithm}
    \SetKwInOut{Input}{Input}
    \SetKwInOut{Output}{Output}

    \underline{function NanoMapQuery} $(\mathbf{x}^\mathcal{B}_{query})$\;
    \Input{body frame query point $\mathbf{x}^\mathcal{B}_{query} = \mathcal{N}(\mathbf{\mu}^\mathcal{B},\Sigma^\mathcal{B}) $}
    \Output{$i$, index of frame containing view \\
    $\mathbf{x}^{\mathcal{S}_{i}}_{query} = \mathcal{N}(\mathbf{\mu}^{\mathcal{S}_{i}}, \Sigma ^{\mathcal{S}_{i}} )$ \\
    $k$-nearest-neighbors $\mathbf{x}^{\mathcal{S}_{i}}_1$, ..., $\mathbf{x}^{\mathcal{S}_{i}}_k$ 
    }
    Transform query point from body frame into most recent sensor frame: $\mathbf{x}^{\mathcal{S}_{0}}_{query} \leftarrow \mathcal{N}(T^{\mathcal{S}_{0}}_\mathcal{B} \mathbf{\mu}^\mathcal{B}, \ \Sigma^{\mathcal{S}_{0}}_\mathcal{B} + R^{\mathcal{S}_{0}}_\mathcal{B}\Sigma^\mathcal{B})$
    \\
    \If{IsInFOV($\mathbf{x}^{\mathcal{S}_{0}}_{query}$)}{return 0, $\mathbf{x}^{\mathcal{S}_{0}}_{query}$, Knn($\mu^{\mathcal{S}_0}$);}
    \For{$i\leftarrow 1$ \KwTo $N$}{
    Transform query point into previous frame:
    $\mathbf{x}^{\mathcal{S}_{i}}_{query} \leftarrow \mathcal{N}(T^{\mathcal{S}_{i}}_{\mathcal{S}_{i-1}}\mathbf{\mu}^{\mathcal{S}_{i-1}}, \ \Sigma^{\mathcal{S}_{i}}_{\mathcal{S}_{i-1}} + R^{\mathcal{S}_{i}}_{\mathcal{S}_{i-1}}\Sigma^{\mathcal{S}_{i-1}})$ \\
    \If{IsInFOV($\mathbf{x}^{\mathcal{S}_{i}}_{query}$)}{return $i$, $\mathbf{x}^{\mathcal{S}_{i}}_{query}$, Knn($\mu^{\mathcal{S}_i}$);}
    }
    return ``out of known space", $\mathbf{x}^{\mathcal{S}_{0}}_{query}$, Knn($\mu^{\mathcal{S}_0}$);
     \caption{NanoMap query algorithm.  Subroutine IsInFOV() is described in Section \ref{isinfov}; Knn() is provided by a single-frame $k$-d-tree query. $N$ is the number of measurements stored in memory.}
     \label{nmknn}
\end{algorithm}

\subsubsection{Uncertainty propagation} Accounting for uncertainty is performed as follows. The query is provided as the mean and covariance of a point in the current body frame $\mathcal{B}$ of the robot $\mathbf{x}^\mathcal{B}_{query} = \mathcal{N}(\mathbf{\mu}^\mathcal{B},\Sigma^\mathcal{B}) $.  The query is first transformed into the frame ${\mathcal{S}}_{0}$ of the most recent sensor measurement, $\mathbf{\mu}^{\mathcal{S}_{0}} = T^{\mathcal{S}_{0}}_\mathcal{B} \mathbf{\mu}^\mathcal{B}$, where $T^{\mathcal{S}_{0}}_\mathcal{B}$ represents the local, relative transform between the current body frame and the recent sensor frame.  $T^{\mathcal{S}_{0}}_\mathcal{B}$ is modeled  with a noisy translation $\mathcal{T}^{\mathcal{S}_{0}}_\mathcal{B}$ with covariance $\Sigma^{\mathcal{S}_{0}}_\mathcal{B}$, and known rotation $R^{\mathcal{S}_{0}}_\mathcal{B}$.  In addition to computational simplification, our choice to model translational uncertainty and not rotational is guided by the practical observation that due to gravity, IMUs provide good observability of roll and pitch, and yaw is only a single integration of a noisy gyrometer (covariance grows $\propto N$ for $N$ measurements), whereas positions are double integration of the accelerometer (covariance grows $\propto N^3$).  Under the assumption of independence between body-frame query point uncertainty and each transform covariance, the variance of the query point in frame ${\mathcal{S}}_{0}$ is simply the sum $\Sigma ^{\mathcal{S}_{0}} = \Sigma^{\mathcal{S}_{0}}_\mathcal{B} + R^{\mathcal{S}_{0}}_\mathcal{B}\Sigma^\mathcal{B}$.  Extending this process to the $i$th sensor coordinate frame, we have
$$\Sigma ^{\mathcal{S}_{0}} = \Sigma^{\mathcal{S}_{0}}_\mathcal{B} + R^{\mathcal{S}_{0}}_\mathcal{B}\Sigma^\mathcal{B}$$
$$\Sigma ^{\mathcal{S}_{i}} = \Sigma^{\mathcal{S}_{i}}_{\mathcal{S}_{i-1}} + R^{\mathcal{S}_{i}}_{\mathcal{S}_{i-1}}\Sigma^{\mathcal{S}_{i-1}} \ \ \text{for } i=1, 2, ... N $$
and concatenating transforms for the mean we have
$$\mathbf{\mu}^{\mathcal{S}_{i}} =  \prod_{j=1}^i \bigg[ T^{\mathcal{S}_{j}}_{\mathcal{S}_{j-1}} \bigg] T^{\mathcal{S}_{0}}_\mathcal{B} \mathbf{\mu}^\mathcal{B}$$
which defines $\mathbf{x}^{\mathcal{S}_{i}}_{query} = \mathcal{N}(\mathbf{\mu}^{\mathcal{S}_{i}}, \Sigma ^{\mathcal{S}_{i}} )$.

\subsubsection{IsInFOV(): determining in-frame views}\label{isinfov}

\begin{figure}
  \includegraphics[keepaspectratio=true,scale=0.60]{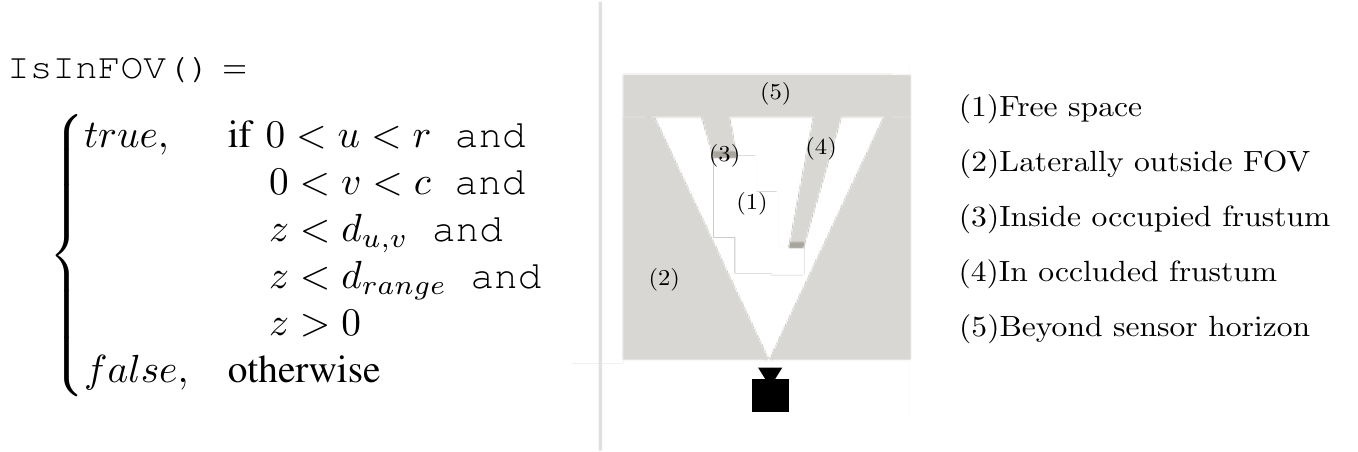}
  \captionof{figure}{The IsInFOV function (left) for determining if a query point is in freespace or one of the regions of non-freespace (right).  The pixel coordinates $u, v$ can easily be calculated by $(u, v) = (\frac{x}{z}, \frac{y}{z})$ and $(x, y, z) = K\mu$ with $K$ the camera intrinsics matrix, and the query point $\mu \in \mathbb{R}^3$ in the right-down-forward Cartesian frame of the sensor measurement.  $r \times c$ is the depth camera resolution.
} \label{isinfov-fig}
\end{figure}

\begin{figure}
  \includegraphics[keepaspectratio=true,scale=0.55]{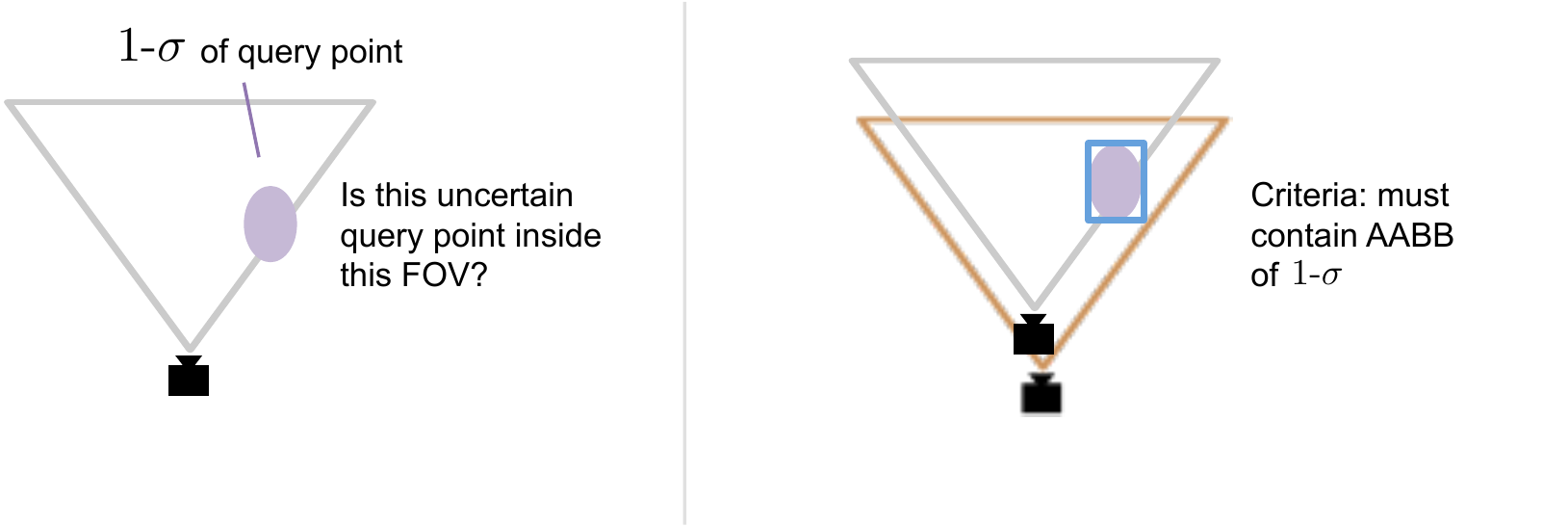}
  \captionof{figure}{Which view is sufficient for evaluating an uncertain query point distributed as an infinite-tail Guassian?  A more recent measurement (left) may contain the mean of the distribution, but a significant portion may fall outside.  Our criteria (right) for an approximate solution is to use the AABB of the 1-$\sigma$ of the query point distribution.  If no view fully contains the AABB, then the most recent frame is used, which may have had a partial view (Algorithm 1).
} \label{whichfov}
\end{figure}

A key challenge is in determining which view contains the uncertain point, referred to as the IsInFOV() function.  Projecting the mean of the uncertain point into the depth image, as described in Figure \ref{isinfov-fig}, can be used to efficiently check a series of inequalities (inside each of lateral and vertical FOV, occluded, not beyond sensor horizon) to determine if the point is in free space.  A challenge, however is represented by Figure \ref{whichfov}.  If only the mean of the distribution is used to check whether or not a view contains the point, then a large portion of that distribution may lie outside the FOV.  With infinite-tail Gaussian distributions, no view fully contains them.  NanoMap approximates this problem by using an axis-aligned bounding box (AABB), a familiar concept for fast approximations in the graphics community.  The AABB for the 1-$\sigma$ (1 standard deviation) of the distribution is used.  Checking whether or not the AABB is contained can be done efficiently with the same number of inequality evaluations as the single point.  To check for occlusions, NanoMap performs a simple occlusion check of the mean point.

\subsection{Data Structure for Asynchronous Data}

\begin{figure}
  \includegraphics[keepaspectratio=true,scale=0.55]{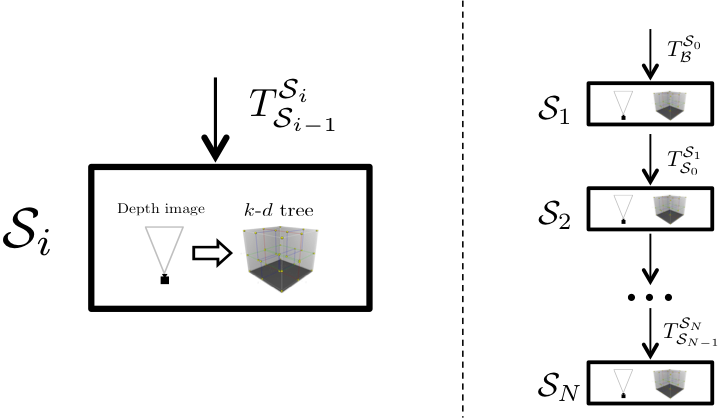}
  \captionof{figure}{Core NanoMap data structure: a sequence (right) of edge-vertex pairs (left), where each vertex contains both the raw data for evaluating FOV constraints, and a $k$-d-tree of the point cloud data.  The edge is a relative transform to the coordinate frame of the next vertex.
} \label{data_structure}
\end{figure}

The data structure (Figure \ref{data_structure}) matches the form of the query algorithm and is performant given the requirements of asynchronous data and continuous addition and removal of data.  The core data structure is a chain of edge-vertex pairs, where the edge is the transform $T^{\mathcal{S}_{i}}_{\mathcal{S}_{i-1}}$ and the vertex contains both the raw point cloud data and the previously-processed $k$-d-tree.  The raw point cloud data (row-column-organized) is used to evaluate the IsInFOV() function, whereas the $k$-d-tree is used to evaluate $k$-nearest-neighbors if IsInFOV()=true.

We briefly highlight some data structure design considerations.  By nature NanoMap is never defined in one coordinate frame, and rather has components in many coordinate frames.  One implication of this is that NanoMap must constantly be updating $T^{\mathcal{S}_{0}}_\mathcal{B}$ with each new pose.  Further, we desired both fast insertion of a new edge-vertex pair, and fast removal of the oldest edge-vertex pair.  Since search through the data structure is also always performed linearly, a doubly-linked list of edge-vertex pairs is a good fit for these requirements, efficiently supporting $O(1)$ addition/removal at ends, and $O(1)$ for each step of IsInFOV().  An additional feature given the separate-frame nature of the framework and asynchronous data is that the $k$-d-tree of a point cloud can be built even before the pose of the point cloud can be determined, allowing the $k$-d-tree building to begin before a world-frame map would be capable of starting insertion.  Lastly, a key feature of NanoMap is to be able to efficiently handle asynchronous updated recent pose information, which may only cover a subset of its history.  Upon receiving a series of updated world-frame poses, NanoMap only updates a transform edge $T^{\mathcal{S}_{i}}_{\mathcal{S}_{i-1}}$ if it can fully interpolate the updated world frame pose of both vertices.  This can be done efficiently by searching through the edge-vertex chain with a time-sequenced list of pose updates.

\begin{table}
    \caption{NanoMap Data Inputs and Parameters}
    \begin{tabularx}{\columnwidth}{X|X|X}
        \hline
        \textbf{Data Input}         & \textbf{Note}   & \textbf{Example Rate} \\ 
        \hline
        6-DOF poses & timestamped & 100 Hz \\ \hline
        6-DOF pose corrections & sequence of timestamped poses & 1-100 Hz        \\ \hline
        Organized 3D PointClouds  & organized (row,column) from depth camera & 30-60 Hz  \\
        \hline
        \textbf{Parameter}         & \textbf{Note}   & \textbf{Example Values} \\ 
        \hline
        Max sensor range & & 10-20 m \\ \hline
        Depth camera resolution, FOV & equivalently, K matrix & 320x240, 60 deg V, 90 deg H FOV \\ \hline
        $N$, history length (\# point clouds) & & 150-300 \\ \hline
         $\Sigma^{\mathcal{S}_{i}}_{\mathcal{S}_{i-1}}$ covariance & between sensor poses & 0.005 - 0.02 m \\ \hline
        
    \end{tabularx}
    \label{table: simulation parameters}
\end{table}


\section{RESULTS}

We start by (A) analyzing in simulation how NanoMap is able to provide robust obstacle avoidance depsite significant state estimation uncertainty, and quantify the scale of drift and correction jumps (i.e., from a loop closure) at which this is significant.  We then (B) analyze the computational efficiency of NanoMap compared to other available packages for evaluating local 3D data in motion planning.  Finally, (C) we demonstrate NanoMap used effectively on a real hardware system.

\subsection{Robustness of NanoMap to State Estimation Uncertainty}\label{robust-results}

A central goal of NanoMap was to increase obstacle avoidance robustness in regimes of significant state estimation uncertainty.  
There are two separate features we evaluate: the ability to separately model pose uncertainty of each depth measurement, and the ability to efficiently correct recent pose information from a sliding-window state estimator.  Our hypothesis was that at some threshold of pose uncertainty, these features become relevant. Here we present our findings.  

\subsubsection{Experimental: Motion Planner and Simulation}

In these experiments, NanoMap is used by a stochastic motion planner.  This motion planner was as described previously \cite{florence16}, with the following modifications: (a) a full 3D motion primitive library of 125 primitives, (b) a collision-chance-constrained (maximum allowed collision probability of $0.001$) rather than mixed-objective described previously, and (c) ``early-exit'' for subsequent sampling of a primitive that already evaluates below the chance constraint.  Our simulation system was also as described in \cite{florence16}, here used with a professional-grade urban environment created in the Unity game engine.  The simulated depth camera was 30 Hz, 20 $m$ range, and 46 $deg$ $FOV_{vertical}$.  

\begin{figure}
  \includegraphics[keepaspectratio=true,scale=0.45]{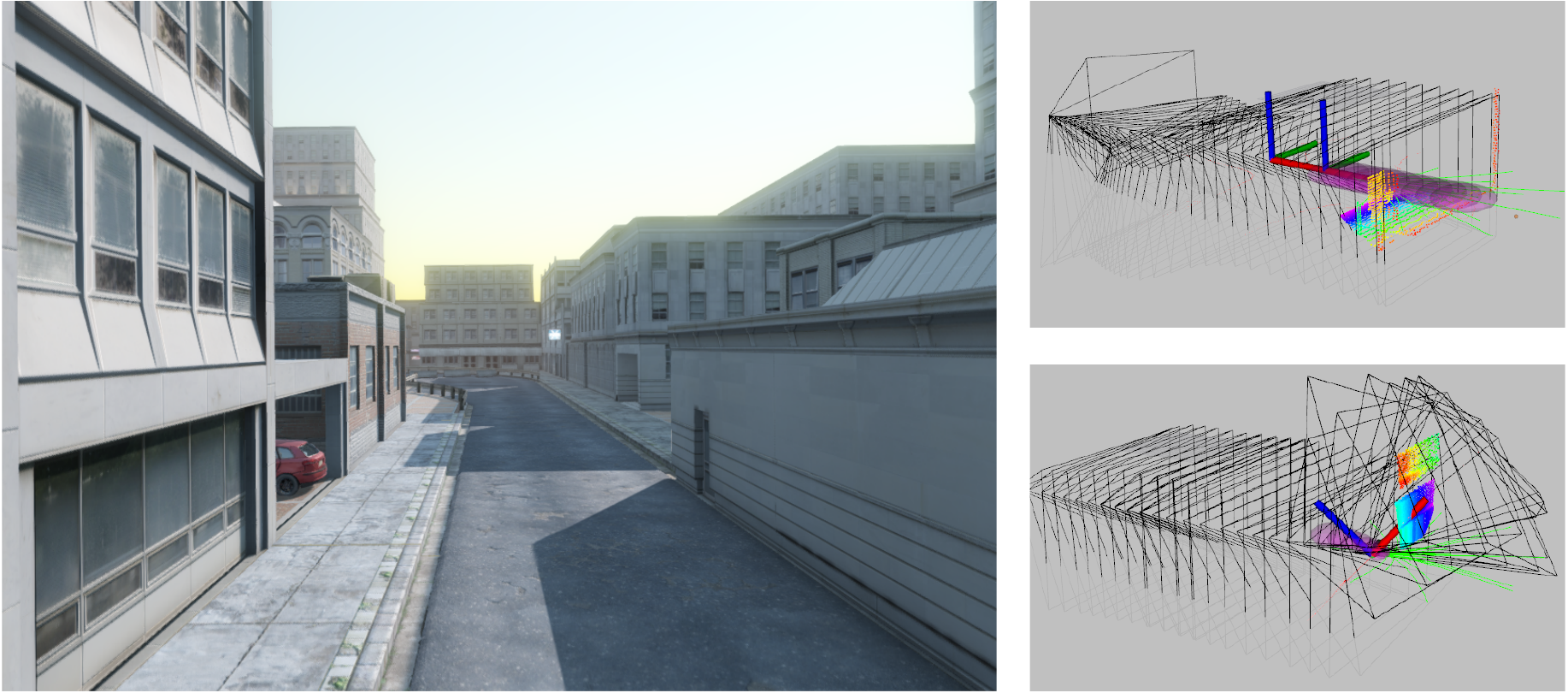}
  \captionof{figure}{Simulation environment (left) at beginning of experimental scenario.  (Right top) Ground truth pose of vehicle and corrupted pose of vehicle (axes are 10 $m$, for scale).  (Right below) Aggressively stopping when far wall comes into depth sensor range.}\label{robustness}
\end{figure}

\subsubsection{Scenario}

The ability of NanoMap to provide pose-uncertainty-aware queries is most relevant when a motion planner is forced to search deeper into its history of poses.  As discussed later with Figure \ref{hist_over_time}, this is most apparent during extreme dodging maneuvers.  Accordingly the experiments use the following scenario which is desirable due to its ease of interpretability: a quadrotor, initially at 5 $m$ altitude, is given a desired goal 200 $m$ away, with a desired top speed of $15$ $m/s$, and 100 $m$ along its path there is a large wall of a building with a 3D overhang near its altitude.  The vehicle must aggressively decelerate, such that its velocity is outside of its current FOV.  Significant pose uncertainty during this aggressive deceleration period would be difficult for other mapping and planning systems to handle.

\subsubsection{Using Pose-Uncertainty-Aware Queries}\label{pose-drift}

To evaluate the magnitude of pose drift at which NanoMap's frame-specific uncertainty capability measurably increases robustness, we experimented with the following controlled experiment.  
As we increased state estimation noise, we either had NanoMap model the local, relative transforms with no translation covariance, $\Sigma^{\mathcal{S}_{i}}_{\mathcal{S}_{i-1}} = \mathbf{0}$, or with a covariance corresponding to the noise level, $\Sigma^{\mathcal{S}_{i}}_{\mathcal{S}_{i-1}} = f(\Sigma_{actual}$).   Our noise model was to add noise to each of the $x$ and $y$ acceleration measurements, $\tilde{a} = (a+\eta) \times \xi$, where $\eta \sim \mathcal{N}(0,\Sigma_{actual})$ and $\xi \sim \mathcal{N}(1,\Sigma_{actual})$. Acceleration noise was integrated into the corrupted velocities and positions.  Since quadrotors can measure altitude directly with downward-facing lidars and barometers, we did not model noise in $z$. An intuitive grasp of the scale of the noise model is best described as the standard deviation of drift over the depth measurement history (5 seconds = 150 measurements at 30 Hz) during the final portion of the flight.  We term this $\sigma_{\text{drift, 5 seconds}}$, and accordingly used 
$f(\Sigma_{actual}) = \frac{\Sigma_{\text{drift, 5 seconds}}}{150}$.  The singular difference between the two groups of the data (Figure \ref{robustness}) was the value of the $\Sigma^{\mathcal{S}_{i}}_{\mathcal{S}_{i-1}}$ parameter in NanoMap.

These experiments show (Figure \ref{robustness}) that incorporating pose uncertainty can have a substantial effect, in particular when the drift is on the order of 10 $cm$ per second. At speeds above 10 $m/s$, this is approximately 1\% position drift, which is comparable to expected performance from our VIO state estimator \cite{steiner2017vision}.  At very small drift ($\sigma_{\text{drift, 5 seconds}} = 0.4 \ m$), there is little noticeable difference, but at $\sigma_{\text{drift, 5 seconds}} = 0.7, \ 1.5, \ 3.8 \ m$, incorporating the uncertainty enables the vehicle to still stay safe 97-98$\%$ of the time, whereas the drift deteriorates the safety of the group that doesn't incorporate pose uncertainty.  The ability to stay safe diminishes at massive levels of drift (7.3 $m$ in 5 seconds), where the pose-uncertainty-modeled group only stays safe 90$\%$ of the time, but still more than the unmodeled group.  The pose-uncertainty-modeled group on average stays much farther away from obstacles, ($\gamma$ = distance to closest obstacle), playing it conservative during the aggressive maneuver.

\begin{figure}
  \includegraphics[keepaspectratio=true,scale=0.47]{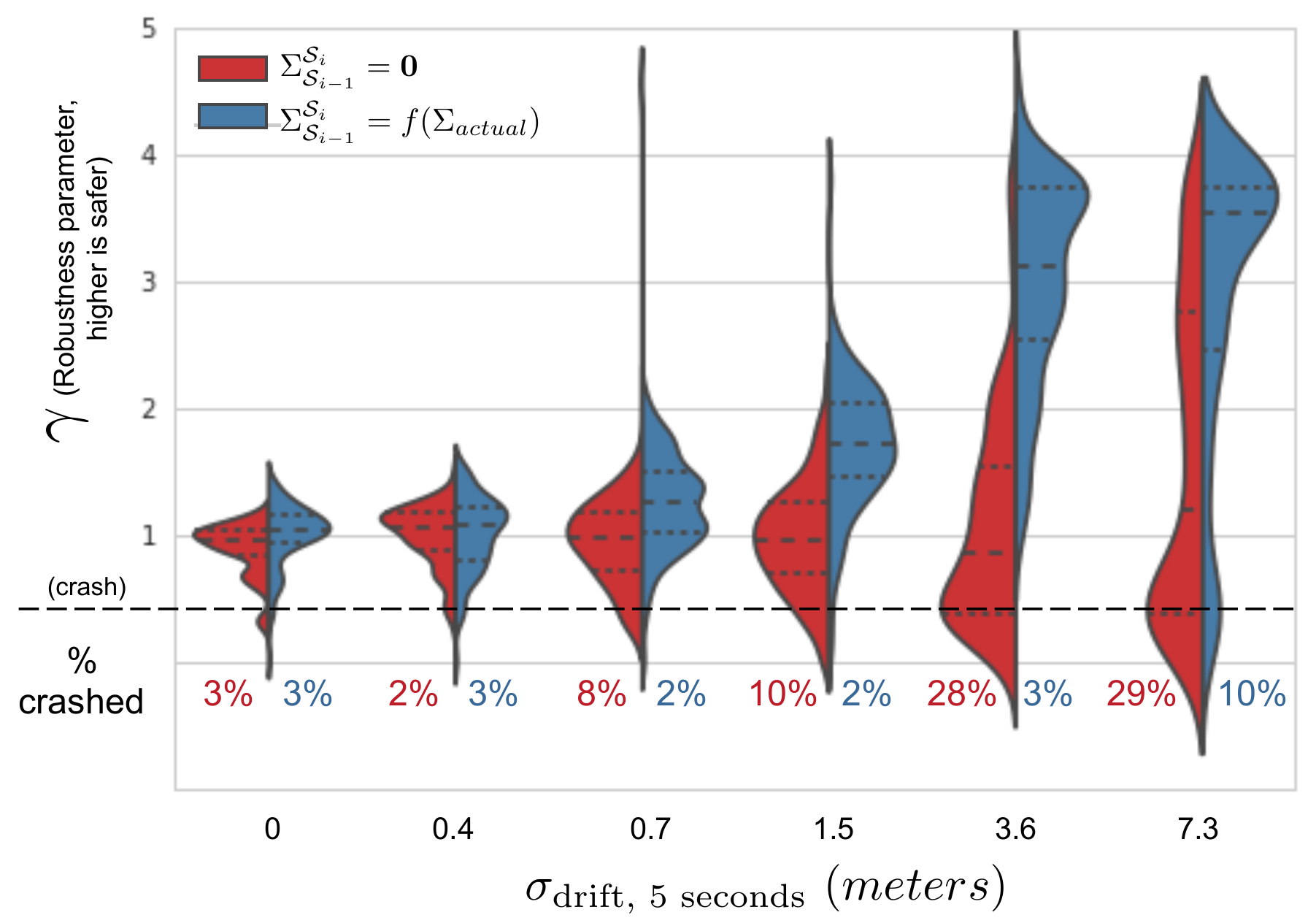}
  \captionof{figure}{Robustness (y axis, where the robustness criteria $\gamma$ is the closest distance in meters to an obstacle during the flight) of either modeling (blue) or not modeling (red) local pose uncertainty.  A sampling of noise levels (x axis) are shown, which represent $\Sigma_{\text{actual}} = \{0, 0.05, 0.1, 0.2, 0.5, 1.0 \}$.  With a vehicle radius of 0.4 $m$, everything below the dotted line represents a crash.  $\%$ crash is labeled for each case.  1200 total trials are represented.}\label{robustness}
\end{figure}

\begin{center}
  \includegraphics[keepaspectratio=true,scale=0.47]{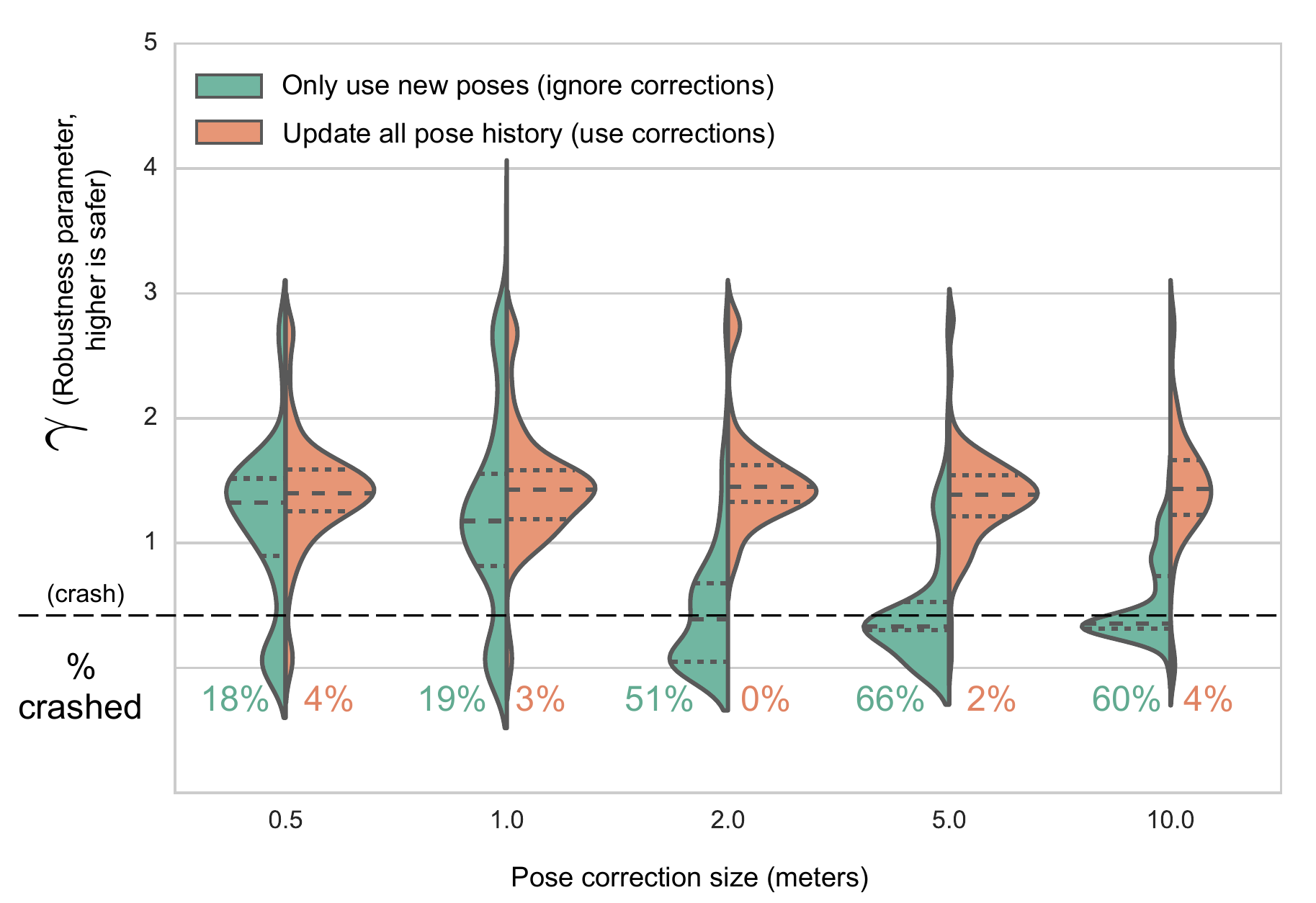}
  \captionof{figure}{Robustness (y axis, where the robustness criteria $\gamma$ is the closest distance in meters to an obstacle during the flight) of either incorporating updated pose history information (orange) after a loop closure or not (green).  A sampling of pose correction sizes (x axis) are shown. As in Figure \ref{robustness}, crash percentages are noted.  1000 total trials are represented.}\label{jump}
\end{center}

\subsubsection{Incorporating Updated Pose Information}\label{pose-jumps}

NanoMap has a unique ability to efficiently update recent pose information, which as shown later in Figure \ref{n_poses}, is not possible at realtime rates for the other benchmarked packages.  This is meaningless, however, without getting a sense of when this capability is useful.  Rather than provide a drifting state estimate, as in the previous experiment, we instead provide a deterministic backwards "pose correction" during the deceleration event (triggered at 12 $m/s$ during the deceleration).  This is representative of a loop closure occurring in the global state estimator.  NanoMap is configured to either use the pose corrections to update and maintain a smooth history of poses, or only add new poses as they come in, and accordingly have a large ``jump'' in its history.

We find that even at the scale of 0.5 $m$ pose corrections, this size of a pose jump in its history can measurably cause crashes during the aggressive maneuver scenario, causing 18$\%$ crashes.  With pose corrections of 2 $m$ or more, these jumps cause crashes more than 50$\%$.  By using NanoMap's capability to trivially update its entire pose history upon receiving a sliding-window correction (orange), there is expectedly little effect for any level of pose jump tested, with crashes occurring less than 5$\%$ at all levels.

\subsection{Computational Efficiency Benchmarking}

We compare NanoMap to three other packages: OctoMap \cite{Hornung:2013:OEP:2458738.2458796}, Voxblox \cite{oleynikova2016voxblox}, and Ewok \cite{usenko2017real}.  OctoMap implements an octree occupancy grid, Voxblox builds ESDFs (euclidean signed distance functions) out of projective TSDFs, and Ewok builds its ESDF by iterating over a 3D circular buffer occupancy grid.  Each of these can provide nearest-obstacle queries, which makes them efficient for stochastic motion planning, where there is uncertainty in configuration. 
There are of course many parameters for each of these packages, but we have made best efforts to provide a useful comparison given reasonable parameter choices.  For both benchmarking experiments, we used a data log of a quadrotor with a simulated 320 $\times$ 240 depth image with 20 $m$ range traversing an approximately $200$ $m \times 200$ $m$ urban environment.  This dataset, and the scripts for using each of these packages to generate the benchmarking data, are available\footnote{https://hub.docker.com/r/flamitdraper/mapping/}.

\begin{figure}
  \includegraphics[keepaspectratio=true,scale=0.40]{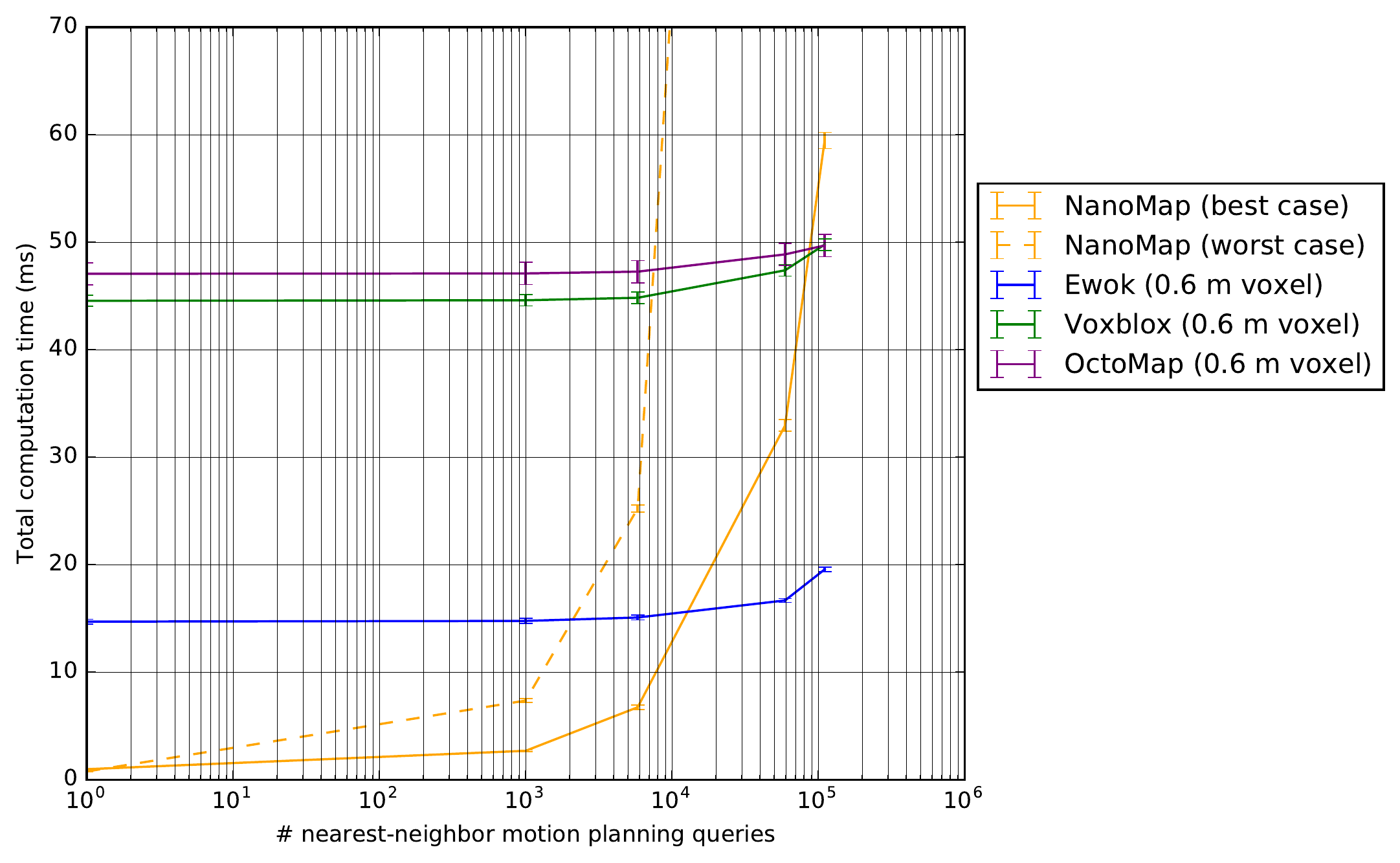}
  \captionof{figure}{Total computation time for $n_{queries}$ nearest-neighbor queries.  NanoMap depth history is set to 150.  Error bars are shown as standard error of the mean.}\label{n_queries}
\end{figure}

\begin{figure}
  \includegraphics[keepaspectratio=true,scale=0.40]{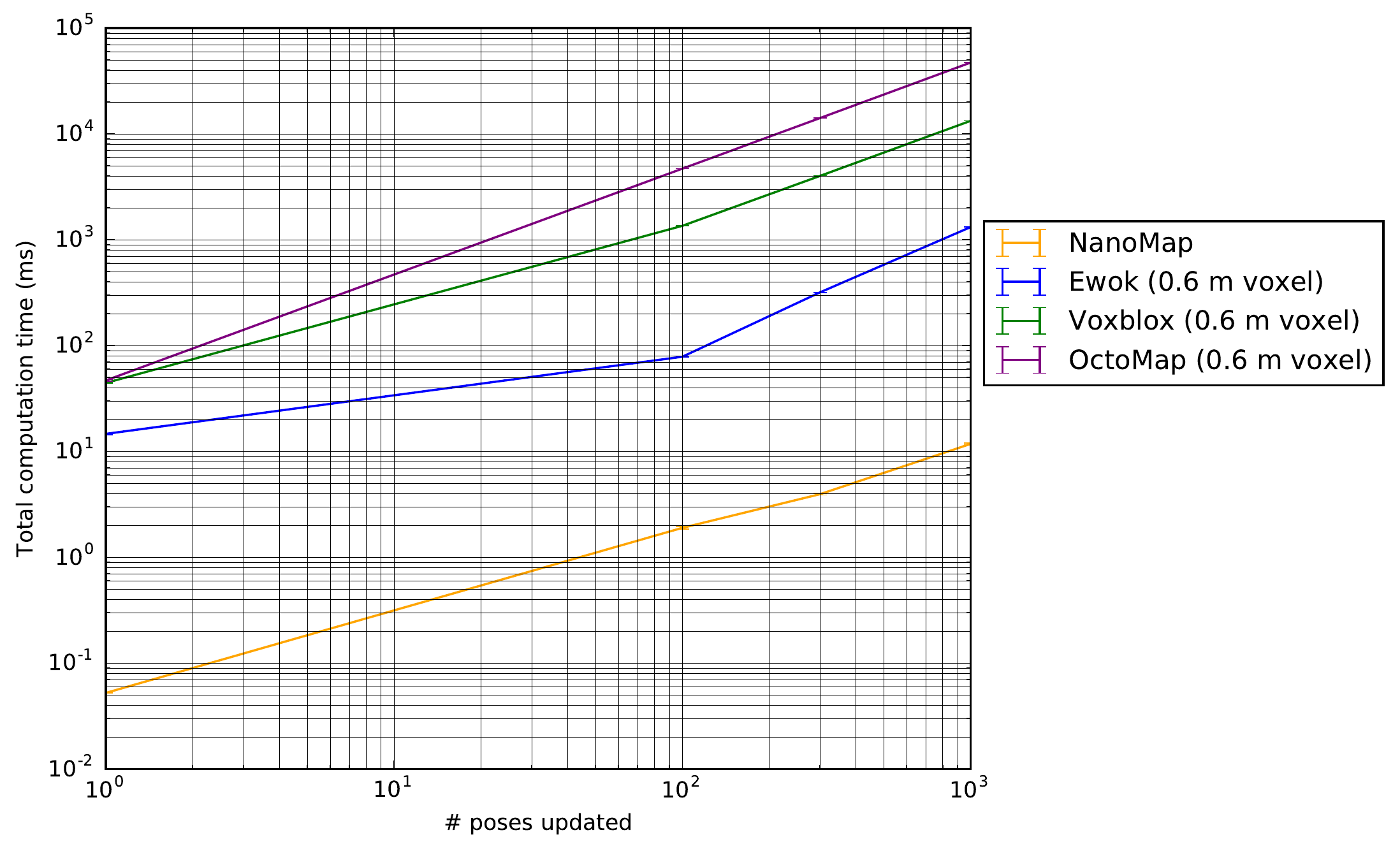}
  \captionof{figure}{Total computation time for adapting map structure to update $n_{poses}$.  Error bars are standard error of the mean.}\label{n_poses}
\end{figure}

We use two metrics to measure the packages.  The first metric (Figure \ref{n_queries}) measures total time to incorporate a new sensor measurement and then perform $n_{queries}$ nearest-obstacle queries.  
The second metric (Figure \ref{n_poses}) measures total time to adjust or rebuild a data structure after $n_{poses}$ poses are corrected, i.e. after a loop closure.

\begin{figure}
  \includegraphics[keepaspectratio=true,scale=0.5]{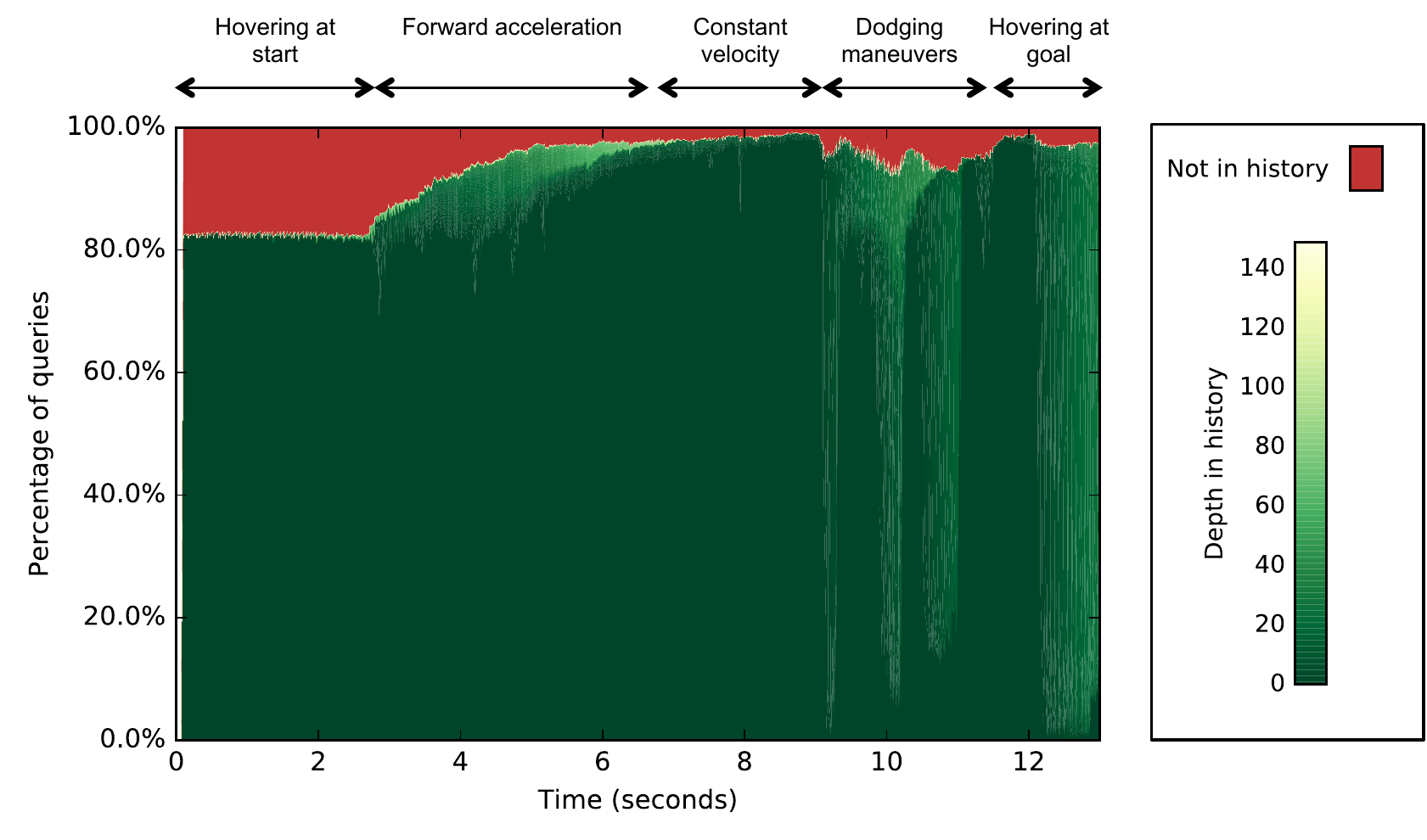}
  \captionof{figure}{Histogram over time for depth of history is searched in a representative flight.  This data is from a 13 second flight traveling approximately 50 $m$ in low clutter with a 30 Hz depth sensor, 10 $m$ range, and 45 $deg$ $FOV_{vertical}$.  Phases of flight are labeled above the time axis.}\label{hist_over_time}
\end{figure}

There are a number of conclusions to draw from the plots.  There is a tradeoff inherent from Figure \ref{n_queries} between the fusion-based packages (OctoMap, Voxblox, Ewok) which spend more time building their data structure, and NanoMap which spends less time building the data structure but has more expensive queries.  For small amounts of queries, this tips the computational advantage to NanoMap, whereas for large amounts of queries, the fusion-based packages have an advantage.  
Figure \ref{n_queries} also demonstrates that unlike the discretized, fused packages, NanoMap has variable query time, based on how deep in history the query searches.  We plot both the worst-case (each query searches the full history) and best-case (each query is in current FOV).  In practice, our planner on average has approximately 75\% best-case queries, but it is important to specify the system to worst-case timing, since as shown in Figure \ref{hist_over_time}, more memory is used during critical dodging maneuvers.  In the range of queries of our motion planner (2,500 queries), NanoMap is the fastest, even in the worst-case.  

From Figure \ref{n_poses}, we see a unique capability of NanoMap -- its ability to incorporate updated pose information at realtime rates.  NanoMap is two to four orders of magnitude faster than the others -- this is not a capability that is feasible at realtime rates for the other packages for more than a handful of $n_{poses}$.  Whereas the only way to incorporate new pose information for the other packages is to rebuild the data structure with new world-frame-registered measurements, NanoMap can adjust by simply updating the relevant sequential transforms ($T^{\mathcal{S}_{i}}_{\mathcal{S}_{i-1}}$) in its data structure.   

We also have been able to empirically validate that for obstacle avoidance motion planning in our flight regimes, a large percentage of NanoMap queries fall within the current or very recent FOV of the depth sensor.  
For a representative flight, we measure a very strong sufficiency of recent measurements, with 74.4\% of queries falling within the current FOV, and a cumulative 92.3\% of queries satisfied with the last 40 measurements.   Figure \ref{hist_over_time} shows a histogram plotted over time during the course of the flight.  During aggressive obstacle avoidance maneuvers (between $\sim$ 9 to 11 seconds into flight), there is expectedly more of a need to use memory.  Yet even during this period, the last few seconds of flight mostly suffice for satisfying motion planning queries. \\

\subsection{Hardware Experimentation}

NanoMap has been extensively used in our hardware system on our MIT-Draper DARPA FLA\footnote{DARPA Fast Lightweight Autonomy program: https://www.darpa.mil/program/fast-lightweight-autonomy} team.  Figure 1, a, shows images from onboard video of a flight.  Over the course of a week of experimental testing at the May 2017 FLA event, NanoMap was the local mapping representation used for the majority of flights, with both an Intel RealSense r200 (for outdoor environments) and an ASUS Xtion (for indoor environments) used as the depth camera sensor.  A Hokuyo 2D lidar sensor also aided obstacle perception for many of these flights, but it was used in a memoryless fashion, and due to its 0-$deg$ vertical FOV was not useful during aggressive high-attitude maneuvers.  We also (see video) demonstrate flight using only the RealSense, with no Hokuyo lidar.  A sliding-window visual inertial (VIO) state estimator \cite{steiner2017vision} with 100 Hz low-latency poses and lower-rate, higher-latency pose corrections over a 5-second sliding window. NanoMap incorporated these sliding window pose corrections. The mapping, planning, and hardware systems have been described in the author's Master's Thesis \cite{FlorenceThesis}.  Notable other vehicle hardware includes: a dual-core Intel NUC i7, a 450 mm Flamewheel DJI frame, and monocular Point Grey Flea3 camera and ADIS 16448 IMU for visual-inertial state estimation.

Our hardware experimentation with NanoMap demonstrates its robustness and applicability to high-speed obstacle avoidance.  We have flown at up to 10 $m/s$ in forested canopy environments with the Intel r200 (empirically, we observe 20+ $m$ range in high-texture environments), and 8 $m/s$ in indoor warehouse environments with the ASUS Xtion (empirically, we observe $\sim$ 8-10 $m$ range).  Flights in these types of settings can be seen in our video.

\section{CONCLUSION}

We have described, implemented, analyzed, and validated NanoMap.  NanoMap provides novel features for using local 3D data with pose uncertainty.  Specifically, it (a) models relative positional uncertainty into its response to local 3D data queries, (b) uses the minimum-uncertainty view to respond to these queries, and (c) can trivially incorporate updated pose information two to four orders of magnitude faster than the benchmarked alternatives.

We have shown that for state estimation drift on the order of tens of $cm/s$ (about 1\% position drift at speeds above 10 $m/s$), or state estimate position corrections on the order of 1 $m$, using NanoMap's uncertainty-aware features can substantially increase robustness.  Given these results, we believe NanoMap is a compelling, novel route forward when compared to the traditional, fusion-first paradigm of mapping for planning.  We would encourage future work that may draw inspiration from NanoMap and supplement traditional mapping approaches.  NanoMap is open source and available at \href{https://github.com/peteflorence/nanomap_ros}{\texttt{github.com/peteflorence/nanomap\_ros}}.

\section*{ACKNOWLEDGMENT}

The authors would like to thank the rest of the MIT-Draper FLA team, particularly Brett Lopez, Kris Frey, Jonathan How, Ted Steiner, William Nicholas Greene, and Nicholas Roy. This work was supported by the DARPA Fast Lightweight Autonomy (FLA)
program, HR0011-15-C-0110.

\bibliographystyle{IEEEtran}
\bibliography{icra-bib}

\end{document}